\documentclass[runningheads]{llncs}
\usepackage{graphicx}
\usepackage{algorithm}
\usepackage{algpseudocode}
\usepackage{amsmath}
\usepackage{xcolor}
\usepackage{hyperref}

\newcommand{\fmo}{$FMO_{M, \phi_{X^*}, x_i}$}
\newcommand{\fmoN}{ $FMO_{M, \phi_{X^*}, x_i}$ }
\newcommand{\itr}{$ITR_{M,R,c}$}
\newcommand{\partialassignment}{$\phi_{X^*}$}

\newcommand{\githubRepo}{\url{https://github.com/autonlab/bn-sat-verification}}

\begin{document}
%
\title{A SAT-based approach to rigorous verification\\ of Bayesian networks}
%

\author{Ignacy Stepka\inst{1}\thanks{Corresponding author.}\orcidID{0009-0004-4575-0689} \and
Nicholas Gisolfi\inst{1}\orcidID{0000-0002-9258-6285} \and
Artur Dubrawski\inst{1}\orcidID{0000-0002-2372-0831}}
\authorrunning{I. Stepka et al.}
%
\institute{Carnegie Mellon University, Pittsburgh PA, USA
\email{\{istepka,ngisolfi,awd\}@andrew.cmu.edu}}


%
\maketitle              
\begin{abstract}


Recent advancements in machine learning have accelerated its widespread adoption across various real-world applications. However, in safety-critical domains, the deployment of machine learning models is riddled with challenges due to their complexity, lack of interpretability, and absence of formal guarantees regarding their behavior. In this paper, we introduce a verification framework tailored for Bayesian networks, designed to address these drawbacks. Our framework comprises two key components: (1) a two-step compilation and encoding scheme that translates Bayesian networks into Boolean logic literals, and (2) formal verification queries that leverage these literals to verify various properties encoded as constraints. Specifically, we introduce two verification queries: if-then rules (ITR) and feature monotonicity (FMO). We benchmark the efficiency of our verification scheme and demonstrate its practical utility in real-world scenarios.


\end{abstract}
\section{Introduction}
In recent years, artificial intelligence (AI) has attracted significant research interest, fueled by its potential to revolutionize various practical applications. Among the many AI models, Bayesian networks (BNs) \cite{darwiche_bayesian_2010_1} stand out in fields that demand extensive expert knowledge. One of the most impactful areas for BNs research is healthcare industry~\cite{mclachlan_bayesian_2020,hill_assessing_2023,vepa_using_2021}. Their adaptive nature, which allows for construction based on either data, expert input, or both~\cite{Druzdzel1995ElicitationOP}, is particularly valuable in incorporating the nuanced expertise of medical professionals. This feature is crucial in healthcare, where understanding the decision-making process can make a profound difference in patient outcomes.

Despite their potential, BNs remain underutilized in real-world clinical practice~\cite{KYRIMI2021102108}. The healthcare sector, characterized by its high stakes and stringent safety requirements, demands absolute reliability and accountability. Even minor errors can have severe consequences, thus placing a serious responsibility on medical practitioners. From our review of recent literature, such as~\cite{KYRIMI2021102108}, we posit that the limited adoption of BNs (and other AI systems) in such critical environments is partly due to a lack of comprehensive understanding of these models’ strengths and weaknesses, what they learned, the reasons for particular decisions, as well as their potential limitations. This situation mirrors challenges faced in other safety-critical industries, such as avionics, where rigorous software certification protocols~\cite{urban2021review_1,formal_methods_aerospace,dmitrievSoftwareDevBasedOnISOavionics} are essential before deployment and have been established for a long time. This provides evidence that implementing similar AI verification schemes could be pivotal in facilitating the integration of AI into complex, high-stakes environments.

To bridge this gap, we propose a formal verification approach for BNs, aimed at enhancing their deployment by ensuring rigorous verifications and sanity checks. This approach seeks to confirm that a given model adheres to critical design specifications, which is especially important in scenarios where errors can be life-threatening. By enabling comprehensive testing for all potential adverse scenarios, our method guarantees that under specified conditions, the model will never execute undesired actions. This not only increases confidence in the model's reliability but also paves the way for broader adoption of AI in critical domains like healthcare.

In this paper, we introduce a novel scheme that first compiles Bayesian networks into Multi-valued Decision Diagrams (MDDs)\cite{shih_compiling_2019_1} and then encodes these diagrams as Boolean algebra formulae, specifically in Conjunctive Normal Form (CNF)\cite{abio_on_CNF}. While previous approaches have compiled BNs into Boolean algebra for inference purposes~\cite{satBNinference}, our method offers two highly desirable properties due to the compilation algorithm: it transforms the probabilistic representation of BNs into Boolean algebra formulae in a bounded time and graph size, and the compiled form is easier to understand, facilitating the development of complex verification queries.

The main contribution of this paper lies in the introduction of formal verification queries, enabling exact verification of desired specifications. We define two novel verification queries for Boolean-encoded Bayesian networks. The first, "if-then rules," verifies whether a premise resulting in a desired outcome is always true for the model. The second, "feature monotonicity," checks if the relationship between a set of facts (i.e., feature assignments) and the outcome variable is monotonic (positive or negative). Unlike existing approximate monotonicity verification methods~\cite{monotonicityBN}, our approach provides exact verification within a SAT formalism. Furthermore, to facilitate model debugging, we define the verification queries using a proof-by-contradiction approach, enabling the enumeration of counterexamples that do not satisfy the query. This allows developers and experts to identify and address specific aspects of the model that fail to meet the verification criteria.

Our approach can significantly enhance the deployment of Bayesian networks in real-world settings by ensuring their adherence to critical design specifications. Additionally, our framework can function as a standalone testing component, performing sanity checks within a larger testing suite to establish a safe and responsible deployment process, akin to accreditation processes in~\cite{Gisolfi2022}.

To summarize, the main contributions of this paper are:
\begin{enumerate}
\item Introduction of an end-to-end scheme utilizing existing compilation~\cite{shih_compiling_2019_1} and encoding~\cite{abio_on_CNF} methods to enable formal verification.
\item Specification of two verification query types and their encoding schemes, allowing experts and developers to verify specific model properties.
\item Development of a counterexample enumeration mechanism to gain insights into model behavior.
\item Release of an open-source implementation for the formal verification of Bayesian networks \githubRepo.
\end{enumerate}

\section{Related Work}

Many safety-critical industries, such as automotive and aerospace, have rigorous testing and certification protocols~\cite{avionicsISO,dmitrievSoftwareDevBasedOnISOavionics} to ensure quality and reliability. However, no such widely accepted framework currently exists for Artificial Intelligence, though several works aim to address this capability gap, such as~\cite{amlas,Gisolfi2022}.

Closely related is the field of software verification, which is well-established and applied across various real-world applications, including the avionics industry~\cite{urban2021review_1}. There are multiple methods for performing formal verification of software, which can be categorized into distinct groups based on varying underlying paradigms. Relevant to our work are the following methods: (1) Deductive Verification~\cite{Floyd1967AssigningMT}, which involves propagating formal specifications through a program and verifying them with symbolic execution using, for example, a SAT solver~\cite{minisat}. (2) Design by Refinement~\cite{abrial_hoare_chapron_1996}, which is based on successive refinement of a solution in a step-by-step manner, with each step containing a formal proof verified by an automated solver. (3) Model Checking~\cite{QueilleCESAR}, which performs exhaustive exploration to automatically verify properties of a given model, providing a sound and complete verification result.

In this paper, we adapt the model checking approach to Bayesian networks to verify their adherence to certain properties, inspired by deductive verification methods. Moreover, our method can be employed as a verification element in a design-by-refinement effort, thanks to our verification queries' proof-by-contradiction approach, which allows for the enumeration of counterexamples that break constraints.

\subsection{AI Verification}
Software verification techniques cannot be immediately applied to AI systems due to the complexity, probabilistic nature and stochastic processes of model fitting. Verification of AI and machine learning models presents new challenges, such as building compact logical representations. Recent efforts have focused primarily on neural networks~\cite{monotony_neural_network,VerifAI,AI2_SafetyRaobustnessCerification_NN,towardsVerifiedAI,lstm_verif}, given their widespread use in numerous applications. However, research has also extended to other model classes, such as Random Forests~\cite{Gisolfi2022} and other tree ensembles~\cite{sato2019formal_decisiontreeensemble}.

In this work, our objective is to expand the field of formal verification of AI to include Bayesian network (BN) models. Specifically, we aim to enable model checking of BNs using the deductive verification paradigm. In recent years, \cite{shih_compiling_2019_1} introduced an approach to Bayesian network compilation, enabling the efficient representation of BNs in the symbolic form of ordered decision diagrams~\cite{miller_construction_2002}. This advancement opens new possibilities for formal verification. Previous works, such as \cite{shih_symbolic_2018,Shih2018FormalVO}, proposed several verification queries executable on the compiled BN. Our proposal goes a step further by encoding the network into Boolean algebra, opening a new avenue for Bayesian network verification. This development is crucial not only for verifying individual networks but also for integrating them into larger systems, aligning with the industry's widespread acceptance of the Boolean algebra formalism.

As mentioned before, the primary contribution of this paper is the introduction of formal verification queries, facilitating the exact verification of specified requirements. Our novel "if-then" verification query represents a new concept, while our "feature monotonicity" query is related to existing proposals for approximate monotonicity verification in Bayesian networks~\cite{monotonicityBN}. However, our approach to monotonicity verification is distinct as it allows for exact verification and is specifically designed for Boolean logic.

\section{Compilation and Encoding Procedure}
In this section, we present a compilation and encoding procedure that transforms a probabilistic model into a Boolean algebra formula. This process consists of the following two consecutive steps: 
\begin{enumerate}
    \item  Utilizing the approach proposed by~\cite{shih_compiling_2019_1}, compile the BN model into a Multivalued Decision Diagram (MDD)~\cite{miller_construction_2002}. The MDD represents the BN's probabilistic decision function in a symbolic and deterministic form. 
    \item Leveraging the scheme from~\cite{abio_on_CNF}, encode the MDD into a Conjunctive Normal Form (CNF) formula, representing the decision function as a Boolean algebra formula. 
\end{enumerate}

\subsection{Bayesian networks}
\label{bayesian-network-section}

\begin{figure}[ht]
    \centering   
    \includegraphics[width=0.4\linewidth]{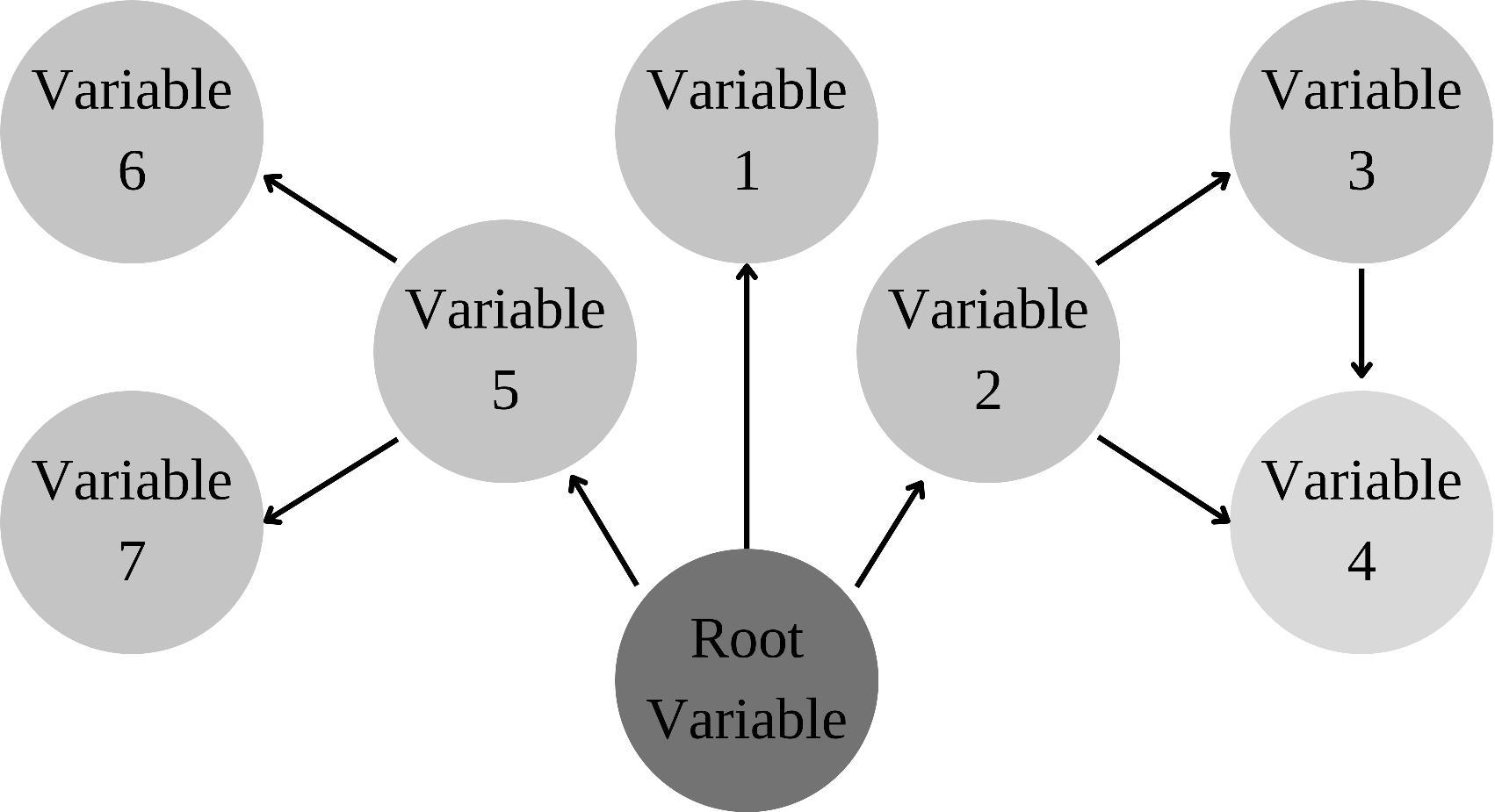}
    \hspace{6pt}
    \includegraphics[width=0.3\linewidth]{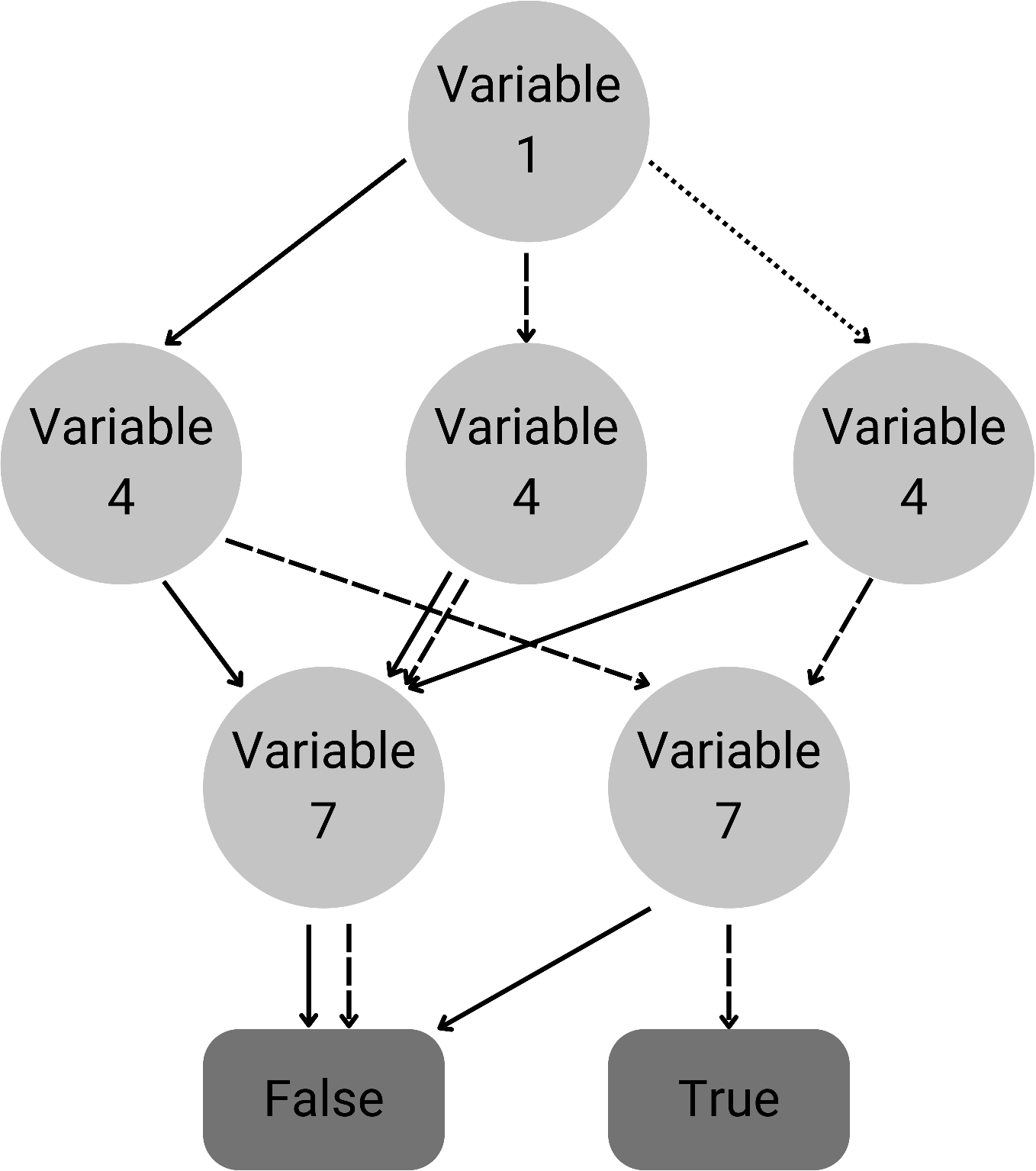}
    \caption{Examples of a Bayesian network (left) and a decision diagram (right).}
    \label{fig:structures}
\end{figure}

Bayesian networks belong to the family of probabilistic graphical models. They are represented by a directed acyclic graph (see an example in Fig.~\ref{fig:structures}) with a set of nodes and edges, that stand for variables and conditional dependencies, respectively. In this work, we assume that there exists one binary class label in the BN, that we will call $Y$. We can represent the decision function as a conditional probability of a class $Y$ given the evidence, $X$ i.e., $P(Y | X_1, X_2, ..., X_n)$. For a more detailed introduction to Bayesian networks, refer to ~\cite{darwiche_bayesian_2010_1}.

\subsection{Compiling BN into MDD}
\label{compiling-bn-into-mdd}
In order to transform the model from a probabilistic representation to a symbolic one, a compilation method is needed. There exist a number of such compilation methods (e.g., \cite{satBNinference,flores_incremental_2011}), however, in this paper we focus on the~\cite{shih_compiling_2019_1} approach which compiles BN into a Multi-valued Decision Diagram (MDD). We motivate this choice by its future potential to extend the compilation algorithm to a multi-class classification scenario, time and space complexity guarantees, easy to use diagram structure, as well as the availability of the open-source implementation. MDDs are concise data structures used to represent functions with multiple-value inputs and binary outputs~\cite{miller_construction_2002}. These tractable and symbolic representations are widely used for optimization purposes in decision-making problems involving complex models. They are built of multiple levels of nodes, where each level can have a various number of nodes, all of which represent the exact same variable, but reached through distinct paths, representing a unique assignment over variables located above it. Terminal nodes in MDDs are called sinks and correspond to True and False outcomes. This representation is a symbolic equivalent of BN's (crisp) decision function, where the predicted class by MDD is the same as the original $Y$ in the BN after applying an arbitrary threshold on the $Y$ probability (that is, typically 0.5).  An example of a MDD is visualized in Fig.~\ref{fig:structures}

\subsection{Encoding MDD into CNF formula}
\label{encoding-into-cnf}

The next step is a transformation of the MDD into a CNF formula using an encoding algorithm adapted from ~\cite{abio_on_CNF}. We choose this particular encoding scheme because it provides many desirable properties accelerating automated theorem proving and results in a compact representation\footnote{These properties are: \textit{consistency}, \textit{domain consistency}, \textit{refutation completeness}, \textit{propagation completeness}. Full definitions can be found in~\cite{abio_on_CNF}}. 
For clarity, we present the exact encoding scheme in Tab.~\ref{tab:decision-diagram-encoding-clauses}. The notation for that table is as follows: $v_i$ is an i-th node in the graph, $\epsilon_{ij}$ is an outgoing edge from $v_i$ to $v_j$, $x_{ij}$ is a variable assignment and $\delta_{ij}$ serves as an incoming edge to $v_i$ from $v_j$.

This encoding scheme mixes two types of approaches. First, it employs Tseitin \cite{Tseitin1983} clauses (T1-T5) aiming to build a consistent representation of the decision diagram in propositional logic. 
Second, it adds path-based clauses (P1-P4) which ensure that the encoding respects the path-based consistency property of decision diagrams (i.e., having at most only one active path at a given time).

\begin{table}[t]
    \caption{Encoding clauses that need to be obtained for every node of decision diagram. Symbol Exactly One (EO) encoding can be obtained by conjunction of At Least One (ALO) and At Most One (AMO). }
    \centering
    \begin{tabular}{l|c|c}
    & Logical formulation & CNF equivalent \\ \hline
    T1: & $v_i \rightarrow \lor_j \epsilon_{ij}$ & $\lnot v_i \lor ALO_j(\epsilon_{ij}) $  \\ \hline
    T2:  & $\epsilon_{ij} \rightarrow v_i$ & $\lnot \epsilon_{ij} \lor v_i$ \\
    T3: & $\epsilon_{ij} \rightarrow \mu_{ij}$ & $\lnot \epsilon_{ij} \lor \mu_{ij}$ \\
    T4: & $\epsilon_{ij} \rightarrow x_ij$ & $\lnot \epsilon_{ij} \lor x_ij$ \\
    T5: & $\mu_{ij} \land x_{ij} \land v_i \rightarrow \epsilon_{ij}$ & $\lnot \mu_{ij} \lor \lnot x_{ij} \lor \lnot v_{i} \lor \epsilon_{ij}$  \\ \hline
    P1: & $v_i \land x_{ij} \rightarrow \epsilon_{ij} $ & $\lnot v_i \lor \lnot x_{ij}  \lor \epsilon_{ij} $ \\
    P2: & $v_i \rightarrow \exists_j \delta_{i-1j} $ where $ v_i \neq \rho$ &  $\lnot v_i \lor_{j} \delta_{i-1j}  $ \\
    P3: & $x_{ij} \rightarrow \exists \epsilon_{ij}$& $\lnot x_{ij} \lor \epsilon_{ij} $\\
    P4: & $EO$($v$ for all $v$ at the same level $i$) & $ ALO(v) \land AMO(v)$
    \end{tabular}
    \label{tab:decision-diagram-encoding-clauses}
\end{table}


In addition to Tab.~\ref{encoding-into-cnf} encoding, we add a synthetic $\rho$ literal to represent a root variable (i.e., $\rho = v_0$) which is always active and ensures satisfiability of the scheme. In contrast to~\cite{abio_on_CNF} we at first omit encoding $T$ and $\lnot F$ (respectively standing for True sink should always be true and False sink should always be false), because later in verification queries, we dynamically add sink encoding as either $T$ and $\lnot F$ or $\lnot T$ and $F$ depending on which class is required to be true.

\section{Verification}
In this section, we introduce two verification queries and the intuition behind them. 
All queries presented in this section are formulated in a proof by contradiction manner, which by default enables enumeration of all \texttt{SAT} models which, in fact, are counterexamples that break the asserted properties of the model. 
Obtaining counterexamples is an important aspect of running verification queries, as it allows decision-makers to empirically examine instances that disprove the model's adherence to the desired specification. 

\subsection{Verification Query \#1: If-Then Rules (ITR)}

The \itr ~query verifies whether a model $M$, for a given set of rules $R$ defined over ordinal variables $X$, will always predict the desired outcome class $c$ for the outcome variable $Y$. The template for defining $R$ is as follows: \textit{if $X_1 \geq t_1$ and $X_2 \geq t_2$ then $Y = c$}, where $t_i$ represents the index of $t_i$-th value in $X_i$. The proposed verification algorithm (see Alg.~\ref{ITR:algo}) constructs a formula that asserts a \textit{conjunction} of all constrained variables. First, each variable is encoded as \textit{disjunction} of all its values satisfying the threshold. Then, the algorithm asserts that the outcome variable $Y$ belongs to any class outside the desired range.   

With that formulation, whenever a \itr verification query evaluates to \texttt{UNSAT}, then $ITR_{M,R,c} = True$. In other words, we can say that the model $M$ adheres to the rule $R$ evaluating in class $c$. Alternatively, when a task evaluates to \texttt{SAT}, we have evidence that an undesired class can be obtained within the asserted range. In simpler terms, this verification query can be intuitively understood as checking for the existence of an unwanted $Y$ class within all feasible hyperrectangles formed under specified constraints over $X$.

\begin{algorithm}[h]
    \caption{If-Then Rules (\itr) Verification}
    \label{ITR:algo}
    \begin{algorithmic}[1]
    \Require M \Comment{Encoded model}
    \Require R \Comment{Constraints on input. In each tuple, the first element is a variable, and the second is a threshold index.}
    \Require c \Comment{Prescribed output (class label)}
    \State $r \gets count(R)$ \Comment{Set r to the number of premises in R}
    \State $F \gets \emptyset$ \Comment{Initialize set $F$ of '\textbf{and}' separated literals}
    \State $CNF_X \gets \emptyset$. \Comment{Initialize set $CNF_X$ of '\textbf{and}' separated literals}
    \State For $i$ from $1$ to $r$: \Comment{Iterate over list of premises}
    
    \begin{enumerate}
        \item $X, t \gets R[i]$
        \item $l \gets card(X)$ \Comment{Assign to $l$ the number of unique variable values}
        \item $C \gets \emptyset$.  \Comment{Initialize set $C$ of '\textbf{or}' separated literals)}
        \item For $j$ from $t$ to $l$: \Comment{Iterate over $X$  variable values above $t$}
        \begin{enumerate}
            \item $C \cup X_j$
        \end{enumerate}
        
        \item $CNF_X \cup C$.
    \end{enumerate}

    \State $F \cup M$ \Comment{Add model}
    \State $F \cup CNF_X$ \Comment{Add correct constraint ranges}
    \State $F \cup Y_{1-c}$ \Comment{Add the outcome of the undesired class}
    \State \itr $\gets$ assert $F$
    \end{algorithmic}
\end{algorithm}

We present a complete pseudocode algorithm for the ITR verification task in Alg.~\ref{ITR:algo}. For simplicity, we assume that all constraints in the verification rules are in the form of $\geq$ sign. However, generalization to all inequality signs is trivial and requires iteration over the respective range in step 4.4 of the algorithm and is included in the source-code.

\subsection{Verification Query \#2: Feature Monotonicity (FMO)}

The \fmo ~query verifies whether a given feature $x_i$ influences the $Y$ variable monotonically, given a partial assignment over some predefined set of other features $\phi_{X^*}$ s.t. $x_i \notin X^*$. In simple words, given a fixed set of facts, it assesses whether a given feature has a monotonic relationship w.r.t. the outcome variable.

\begin{definition}[Partial assignment]
    Partial assignment $\phi_{X^*}$ is a set of variable assignments over a subset $X*$ of variables from $X$. For example, for a set of variables $X*= \{x_1, x_2, x_3\}$,  a partial assignment $\phi_{X^*}$ assigns exemplar values to variables in $X*$ in the following way: $\phi_{X^*} = (x_1 = 0, x_2 = 3, x_7 = 1)$.
\end{definition}

\begin{definition}[Positive Monotonic]
    We say that the prediction function $f: X \rightarrow Y$ is positive monotonic whenever, given increasing values of the variable $x$, the assignments over $Y$ are non-decreasing. 
\end{definition}

\begin{definition}[Negative Monotonic]
    We say that the prediction function $f: X \rightarrow Y$ is negative monotonic whenever, given increasing values of the variable $x$, the assignments over $Y$ are non-increasing. 
\end{definition}

\begin{definition}[Feature Monotonicity]
   We say that a model $M$ is feature monotonic (\fmo), whenever given a $\phi_{X^*}$, the $M$'s prediction function $f$ is both positive and negative monotonic w.r.t. variable $x_i$.   
\end{definition}

\begin{figure}[h!]
    \centering   
    \includegraphics[width=\linewidth]{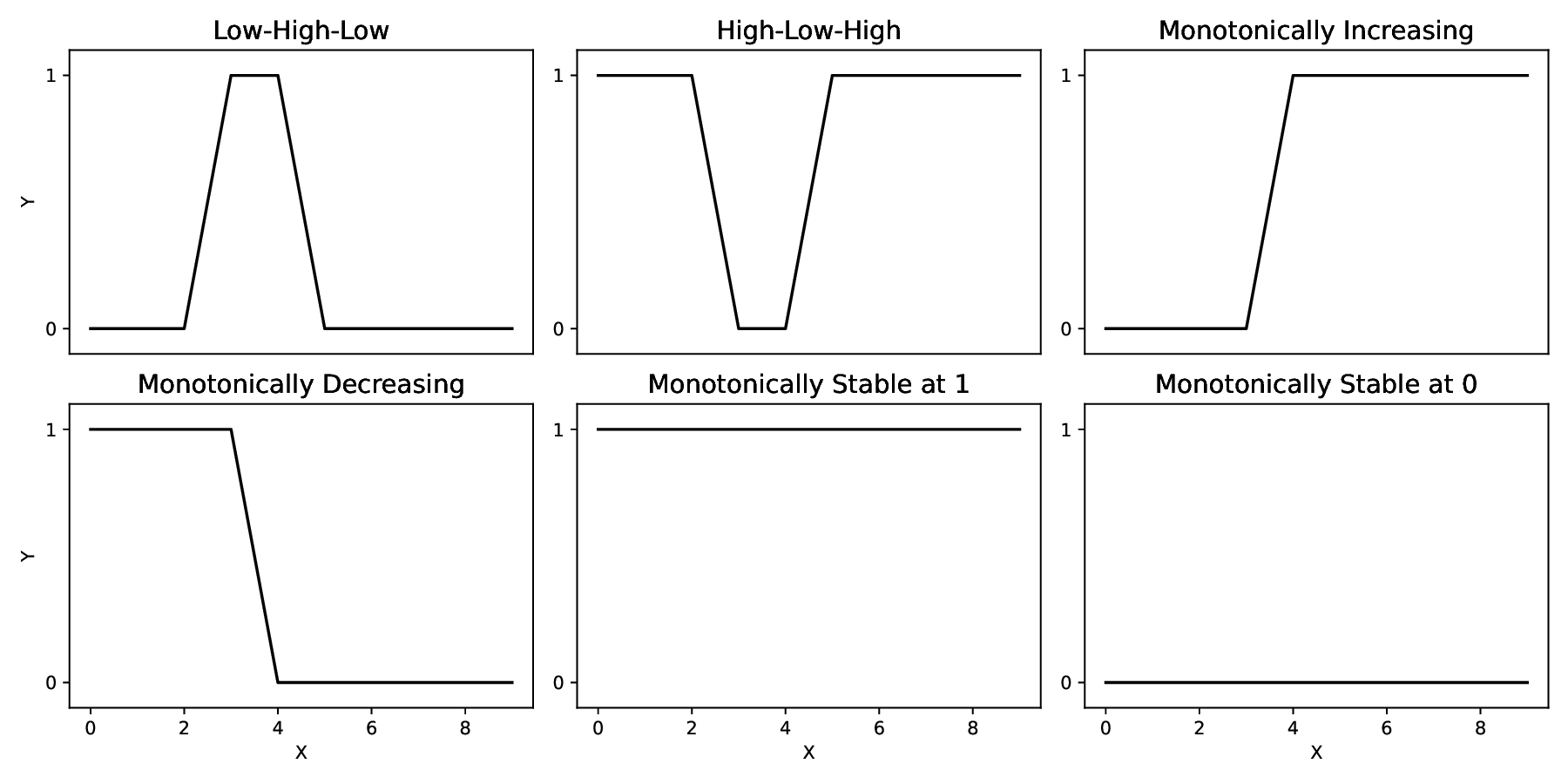}
    \caption{Examples of functions with four different types of feature monotonicity, and two (LHL/HLH) non-monotonic alternatives.}
    \label{fig:fmovisual}
\end{figure}

In this paper, we focus on a binary classification scenario, however, both introduced verification queries support a multi-class classification scenario. Therefore, in a binary setting, \fmo checks whether with gradually increasing values of $x_i$ the predicted class flips at most once. To give more intuition, for example, in a loan approval scenario, this query allows checking whether having increasingly higher credit score will result in an approval decision going from negative to positive with only one flip of the decision along the entire credit score range. In order to define \fmo in a proof-by-contradiction manner, it is essential to assert that an undesired relationship exists along the entire domain of $x_i$. This can be done via assertion of existence of patterns that do not follow the desired monotonicity. To cover both positive and negative monotonic relationships, if either pattern of prediction changes (on the three consecutive values in the $x_i$ domain is in place): \textit{High-Low-High (HLH)} or \textit{Low-High-Low (LHL)} is in place, then the \fmo is violated. Examples of monotonic relationships are presented in Fig.~\ref{fig:fmovisual}, where two top-left plots visualize the LHL and HLH functions, while the other four plots show monotonic functions. In essence, LHL and HLH assert local non-monotonic relationship, which is enough to establish a function as non-monotonic in the domain.

More formally, the proposed algorithm (Alg.~\ref{MO-algo}) creates three copies $M1$, $M2$, $M3$ of the model $M$ and adds constraints that enforce an increasing assignment of the feature $x_i$ in these models in the following way: $\alpha = x_i^{M1} < x_i^{M2} < x_i^{M3}$. Then, it adds constraints that break the monotonicity by enforcing a non-monotonic assignment over class variables between these models $\beta_{HLH} = y^{M2} < y^{M3}, y^{M1}$, where $y$ corresponds to the output class for a given model. To account for all possibilities, it is essential to also test for a polar opposite scenario of \textit{Low-High-Low (LHL)}, which requires only a flip of the sign $\beta_{HLH}$, that is, $\beta_{LHL} = y^{M2} > y^{M3}, y^{M1}$.

\begin{algorithm}[t]
    \caption{Feature Monotonicity (\fmo) Verification}
    \label{MO-algo}
    \begin{algorithmic}[1]
        \Require $M$ \Comment{Encoded model}
        \Require $\phi_X$ \Comment{Partial assignment}
        \Require $x_i$ \Comment{Feature to check the monotonicity on}
        \State Create three copies of $M$: $M1$, $M2$, $M3$
        \State $T \gets \emptyset$ \Comment{Create an empty CNF formula (operator \textbf{and} between elements)}
        \State For $t$ from $1$ to $2$:
        \begin{enumerate}
            \item $F \gets \emptyset$ \Comment{Create an empty CNF formula (operator \textbf{and} between elements)}
            \item $F \cup M1 \cup M2 \cup M3 $  \Comment{Add models' literals}
            \item $F \cup \phi_X$ \Comment{Add partial assignment over all variables}
            \item $F \cup (i_{x_i}^{M1} < i_{x_i}^{M2})$ \Comment{Add inceasing assignment order on $x_i$ in adjacent models}
            \item $F \cup (i_{x_i}^{M2} < i_{x_i}^{M3})$ 
            \item If $t = 1$ then:
                \begin{enumerate}
                    \item  $F \cup (Y^{M2} > Y^{M1})$ \Comment{Add outcome $\beta_{LHL}$ }
                    \item  $F \cup (Y^{M2} > Y^{M3})$
                \end{enumerate}
             else:
                \begin{enumerate}
                    \item  $F \cup (Y^{M2} < Y^{M1})$ \Comment{Add outcome $\beta_{HLH}$ }
                    \item  $F \cup (Y^{M2} < Y^{M3})$ 
                \end{enumerate}
            \item $\tau \gets$ assert $F$ \Comment{Assert the entire formula and return true or false}
            \item $T \cup \lnot\tau$ \Comment{Add negation of the verification result}
            
        \end{enumerate}
        \State \fmo $ \gets$ assert $T$ \Comment{Get the final result of the verification query}

    \end{algorithmic}
\end{algorithm}

Obtaining a CNF encoding of the $>$ inequality between two variables (see lines 6-9 in Alg.~\ref{MO-algo}) can be achieved via encoding the pairwise upper triangle (alternatively, for $<$, the lower triangle) of the 2D Cartesian product matrix of their possible feature values. In this matrix, each cell is a \textit{conjunction} of coordinates $xy$ and all cells are connected by \textit{disjunction}. This approach produces a formula in \textit{ Disjunctive Normal Form} $(DNF)$ which, using Tseitin transformation \cite{Tseitin1983}, is translated from DNF representation to CNF in polynomial time. 

The outcome of $\beta_{HLH}$ and $\beta_{LHL}$ is interpreted jointly, therefore, if either of them is satisfied (i.e., evaluate to \texttt{SAT}): $\beta_{HLH} \lor \beta_{LHL}$, the model is not monotone  w.r.t. $X$ (recall that the proof is done by contradiction, i.e., if constraints can be satisfied, the model does not adhere to verified property). When the formula evaluates to \texttt{SAT}, it allows for enumeration of counterexamples, which are feature assignments that violate monotonicity.  However, when both queries cannot be satisfied (i.e., evaluate to \texttt{UNSAT}): $\lnot\beta_{HLH} \land \lnot\beta_{LHL}$, monotonicity w.r.t. $X$ is guaranteed, because the solver was not able to find any valid assignment that would satisfy all constraints.

\section{Experiments}

In this section, we present two experiments that showcase the utility of the proposed verification framework. In Sec.~\ref{exp:effic}, we present a runtime efficiency benchmark concerning different parts of the framework on a few Bayesian networks from literature. Next, in Sec.~\ref{exp:casestud}, we walk through a comprehensive case study to show the real-world use case of our verification framework in a loan approval scenario. The code used for the experiments is publicly available in an online repository\footnote{\githubRepo}.

\subsection{Benchmarking on Publicly Available Bayesian networks}
\label{exp:effic}

We compiled and encoded five different Bayesian network models from a popular online repository\footnote{\url{https://www.bnlearn.com/bnrepository/}}. Subsequently, we executed verification queries on their encoded representations and recorded the respective runtimes. 
Given the exponential complexity of the compilation algorithm~\cite{shih_compiling_2019_1}, the results presented in Tab.~\ref{tab:dataset-performance} confirm the substantial compilation time increase given larger Bayesian networks. However, it is worth to mention, that this is only an initial cost, which is required to obtain an encoded representation of the model. After that step, any number of verification queries can be executed in a timely manner. 

Next, we verify time efficiency of verification queries. Verification tasks are solved using Minisat SAT solver~\cite{minisat}. While SAT solvers have exponential complexity, they are very capable of swiftly handling large numbers of clauses. As experiments show, the time of solving a single verification task is measured in milliseconds for small BNs, up to low number of seconds for larger ones. 

We contend that such performance properties enable construction of a comprehensive verification suite using the proposed framework. Although there is a relatively high initial entry time cost for compilation, running multiple verification queries becomes rapid after this process, as the entire verification effort requires only one compilation for a given model.

\begin{table}[h]
    \caption{Runtime benchmark statistics averaged over 10 runs on a single CPU. The first 7 rows detail the size of the original BN, the compiled MDD and the encoded CNF. The "Compile [s]" and "Encode [s]" rows indicate the time taken for the respective steps, reported in seconds. The "Verification [s]" row provides an average across 20 randomly initialized \itr and \fmo, with 10 queries each.  }
    \smallskip
    \centering
    \begin{tabular}{l|cccccc}
    Model & \textbf{admission} & \textbf{asia} & \textbf{child}  & \textbf{alarm} & \textbf{win95pts} \\ \hline
        BN - Nodes                  &  3    &   8   & 20  & 37   &  76   \\
        BN - Average node degree         &  1.33 &   2   & 2.5 & 2.49 &  2.95 \\
        BN - \# of parameters   &  10   &   18  & 230 & 509  &  574  \\
        MDD - height             &  3    &   3   & 8   & 12   &   17   \\
        MDD - nodes              &  3     & 3    & 41  & 142  & 260\\
        CNF - literals & 15 & 15 & 210 & 637 & 814 \\
        CNF - clauses & 50 & 50 & 1236 & 4316 & 6257 \\
        \hline
Compile [$s$]            & $2.0 $ & $2.0 $ & $4.5$ & $2.2 \times 10^2$ & $2.8 \times 10^2$ \\
Encode [$s$]             & $5.0 \times 10^{-4}$ & $5.0 \times 10^{-4}$ & $3.6 \times 10^{-3}$ & $1.4 \times 10^{-2}$ & $2.0 \times 10^{-2}$ \\
Verification [$s$]       & $3.0 \times 10^{-4}$ & $3.0 \times 10^{-4}$ & $1.8 \times 10^{-2}$ & $6.6 $ & $3.4 $ \\

    \end{tabular}
    \label{tab:dataset-performance}
\end{table}

Our experiments show that our SAT formalism is able to be applied to existing Bayesian network models without needing to train new models from scratch.
This means that our methods can serve as a reliability assurance layer when connected to existing or legacy systems that implement Bayesian networks.

\subsection{Empirical utility study}
\label{exp:casestud}
In this experiment, we present a study exemplifying the utility of the proposed framework in a verification and validation scenario. Imagine a scenario, where a bank wants to deploy a Bayesian network model to assess whether a loan applicant should be approved for a loan. Since banking industry is highly regulated, the bank would have to meet certain regulatory requirements and prove that their model adheres to required norms and specifications, in order to deploy that model. Below, we simulate this environment and show the practical utility of the proposed framework. 

\paragraph{\bf Data}
For experiments, we selected a dataset from \cite{SMILE_GENIE}, namely Credit10k. It contains 10 thousand samples of credit data organized into 12 discrete columns. We split the data to two train and test subsets, having a 7:3 size ratio.

\paragraph{\bf Bayesian networks}
\label{exper-baynet}
We employ the pgmpy library \cite{ankan2015pgmpy} to train 10 Bayesian networks, with the $CreditWorthiness$ variable as the root node having outgoing edges only. The training process involves two steps: structure learning and parameter estimation. For structure learning, 10\% of the training data is randomly sampled for each task, and a Tree Search estimator (chosen randomly between tan and chow-liu) is used. Subsequently, model parameters are estimated using the entire training set, with the selection of either Maximum Likelihood or Bayesian estimator chosen at random. In the final step, we perform inference on the test data and calculate accuracy to facilitate later analysis.
 

\begin{table}[h]
    \centering
    \caption{Expert rules for \itr verification query. CreditWorthiness is the outcome variable.}
    \label{tab:expert-rules}
    \begin{tabular}{l|c|c|c|c|c|c|c|c|c|c|c|c}
     & rule0 & rule1 & rule2 & rule3 & rule4 & rule5 & rule6 & rule7 & rule8 & rule9 & rule10 \\
    \hline
    PaymentHistory & - & $\leq$ 2 & $\leq$ 2 & 0 & - & - & $\leq$ 1 & 2 & $>$ 0 & $\leq$ 2 & - \\
    WorkHistory & - & $>$ 1 & - & - & - & - & $>$ 2 & - & - & - & - \\
    Reliability & $>$ 0 & $>$ 0 & $>$ 0 & $>$ 0 & $>$ 0 & $>$ 0 & $>$ 0 & $>$ 0 & $>$ 0 & $>$ 0 & $>$ 0 \\
    Debit & $>$ 0 & $>$ 0 & $>$ 0 & - & 0 & $>$ 1 & - & - & 1 & 1 & 0 \\
    Income & - & $>$ 0 & - & $>$ 0 & - & - & - & - & $>$ 0 & - & - \\
    RatioDebInc & $>$ 0 & $>$ 0 & $>$ 0 & $>$ 0 & $>$ 0 & $>$ 0 & 0 & $>$ 0 & $>$ 0 & $>$ 0 & $>$ 0 \\
    Assets & $\leq$ 1 & - & $>$ 1 & - & 0 & - & - & $>$ 1 & - & - & 0 \\
    Worth & - & $>$ 1 & - & - & $\leq$ 1 & - & $>$ 1 & $>$ 1 & - & - & $\leq$ 1 \\
    Profession & 0 & 0 & 0 & 0 & 0 & 0 & 0 & 0 & 0 & 0 & 0 \\
    FutureIncome & $>$ 0 & $>$ 0 & $>$ 0 & $>$ 0 & $>$ 0 & $>$ 0 & $>$ 0 & $>$ 0 & $>$ 0 & $>$ 0 & $>$ 0 \\
    Age & $\leq$ 1 & $\leq$ 1 & $\leq$ 1 & 1 & $\leq$ 1 & 0 & $\leq$ 1 & $>$ 1 & 1 & 0 & $\leq$ 1 \\ \hline
    CreditWorthiness & 1 & 1 & 1 & 1 & 1 & 1 & 1 & 1 & 1 & 1 & 1 \\
    \end{tabular}
\end{table}
\begin{table}[h]
    \caption{Verification experiment. For each model (columns) there is a count of the number of \texttt{SAT} and \texttt{UNSAT} evaluated \itr ~queries (rows).}
    \label{tab:rules_verif}
    \centering
\begin{tabular}{l|cccccccccc}
 & BN0 & BN1 & BN2 & BN3 & BN4 & BN5 & BN6 & BN7 & BN8 & BN9 \\ \hline
\texttt{UNSAT} & 0 & 1 & 0 & 0 & 7 & 2 & 0 & 3 & 0 & 4 \\
\texttt{SAT} & 11 & 10 & 11 & 11 & 4 & 9 & 11 & 8 & 11 & 7 \\
Compliance \% & 0.00 & 9.09 & 0.00 & 0.00 & 63.64 & 18.18 & 0.00 & 27.27 & 0.00 & 36.36 \\
Test Accuracy \% & 70.03 & 72.43 & 70.03 & 68.10 & 72.30 & 72.33 & 72.37 & 72.27 & 72.40 & 68.10 \\
\end{tabular}

\end{table}

\paragraph{\bf Verifying \itr}

To conduct verification queries, we first define a set of expert rules for compliance assessment. For this experiment, we derived 11 rules, as detailed in Tab.\ref{tab:expert-rules}. Subsequently, we performed \itr verification on all involved Bayesian networks. The quantitative results are summarized in Tab.\ref{tab:rules_verif}.

The results indicate that some models have learned decision functions that adhere to the specified requirements more closely than others. Notably, $BN4$ complies with the majority of the rules (7 out of 11). Conversely, several models fail to exhibit the expected behavior. Interestingly, higher compliance with expert rules does not correlate with reduced accuracy, suggesting that adherence to these rules does not necessarily compromise performance.

\begin{figure}[htbp]
    \centering
    \textbf{Query}
    \includegraphics[width=\textwidth]{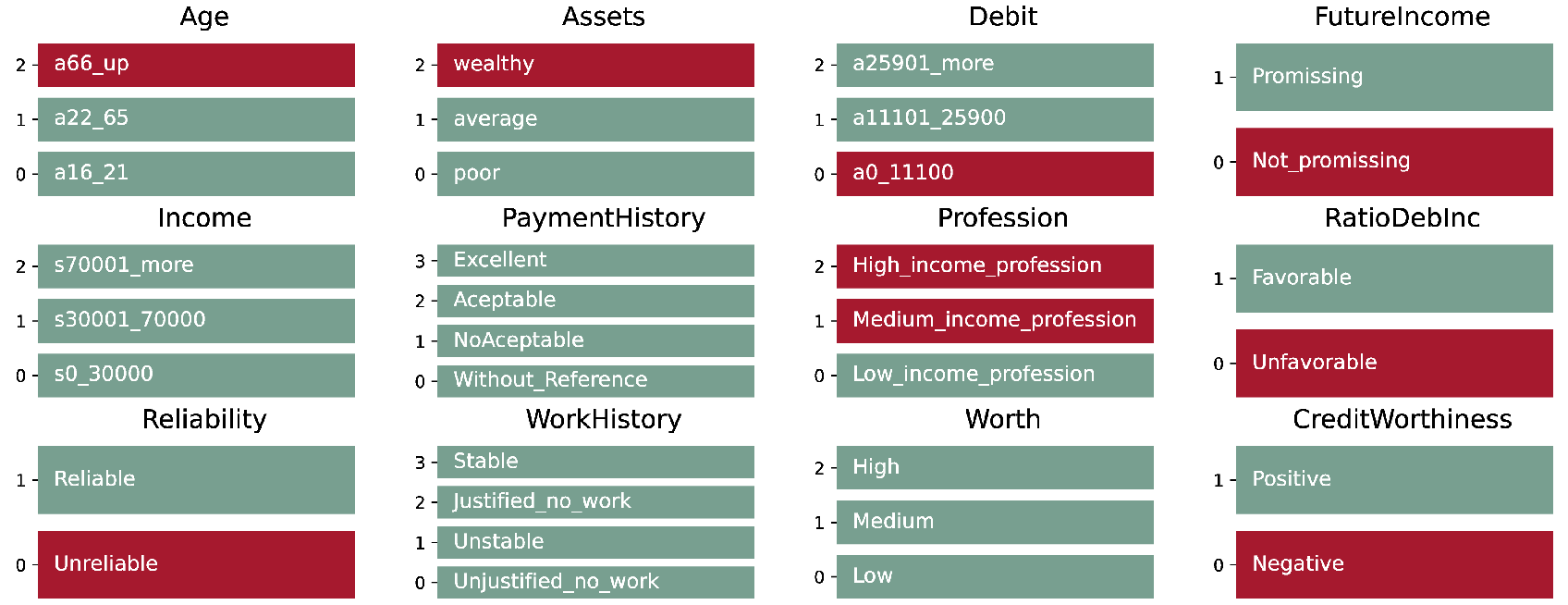}
    \caption{\itr~\texttt{UNSAT} verification query result. Feature values indicated in green are permissible within the constraints, while red bars denote feature values that are not allowed by the constraints.}
    \label{fig:ITR_UNSAT}
\end{figure}

To illustrate the verification capability, we analyze an example of \texttt{UNSAT} and \texttt{SAT} queries for the $BN4$ model. Fig.\ref{fig:ITR_UNSAT} visualizes an \itr evaluation resulting in \texttt{UNSAT} (i.e., adherence to specification) where $R$ and $c$ are set to the values of rule0 from Tab.\ref{tab:expert-rules}. The query's interpretation is straightforward: the model will never alter the predicted class, in this case, from $Positive$ to $Negative$, given the constraints of $R$ (with permitted values shown in green).

\begin{figure}[htbp]
    \centering
        \textbf{Query}
        \smallskip 
        
        \includegraphics[width=\linewidth]{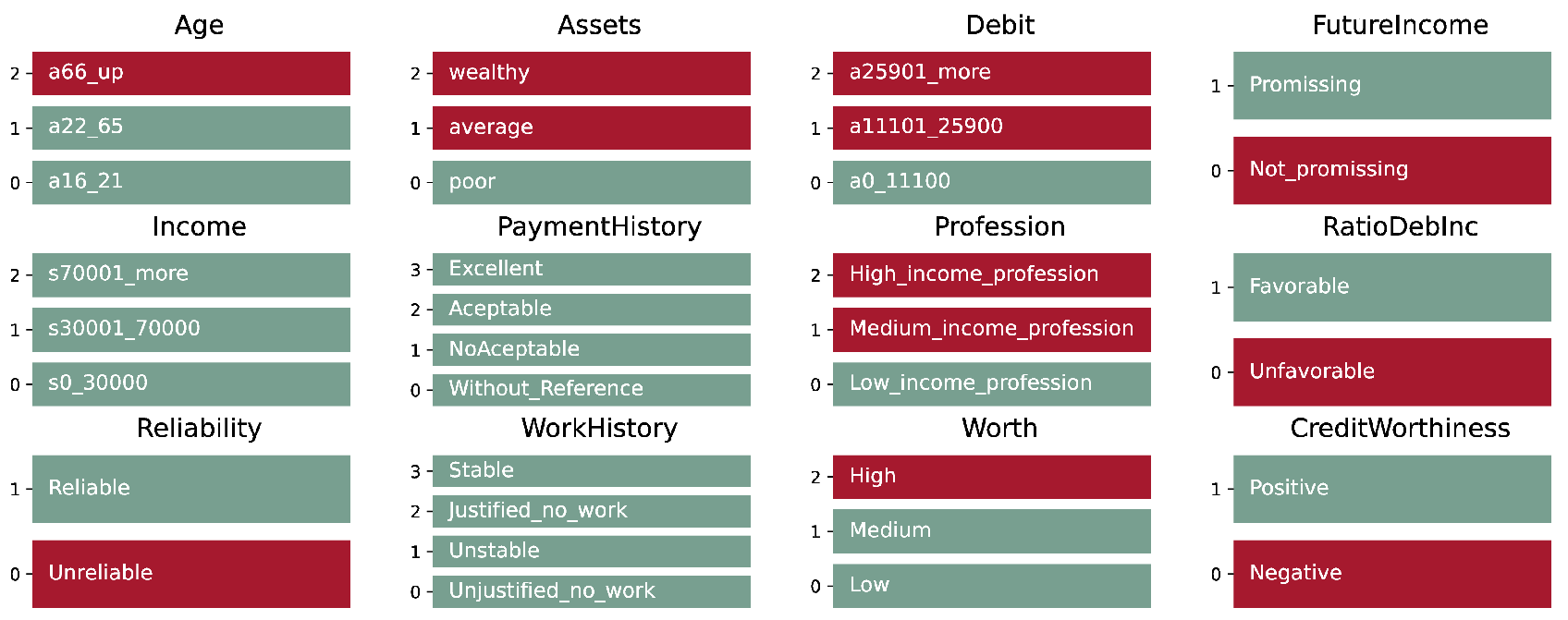} 

        \textbf{Counterexamples}
        \smallskip
        
        \includegraphics[width=0.45\linewidth]{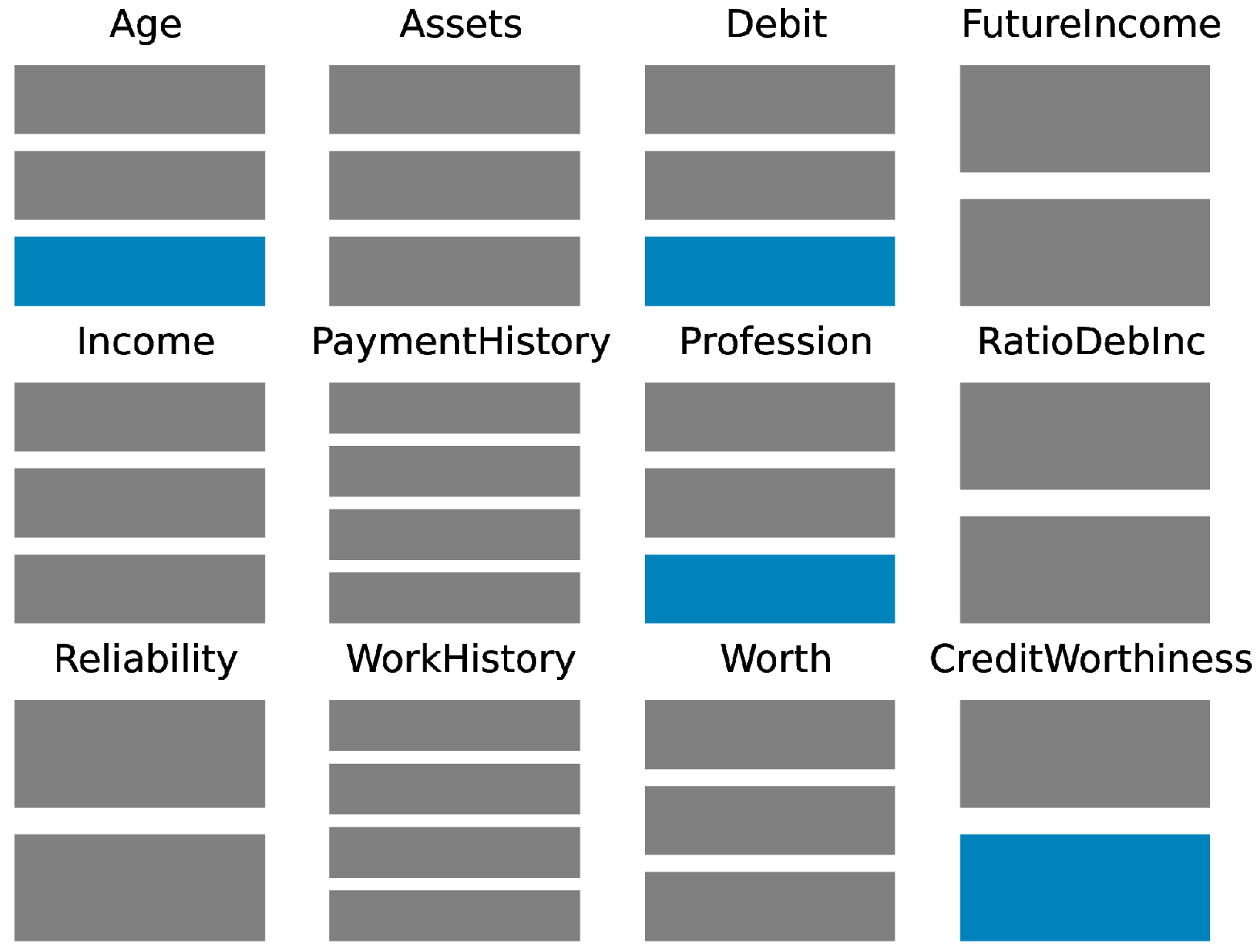} 
        \hspace{10pt}
        \includegraphics[width=0.45\linewidth]{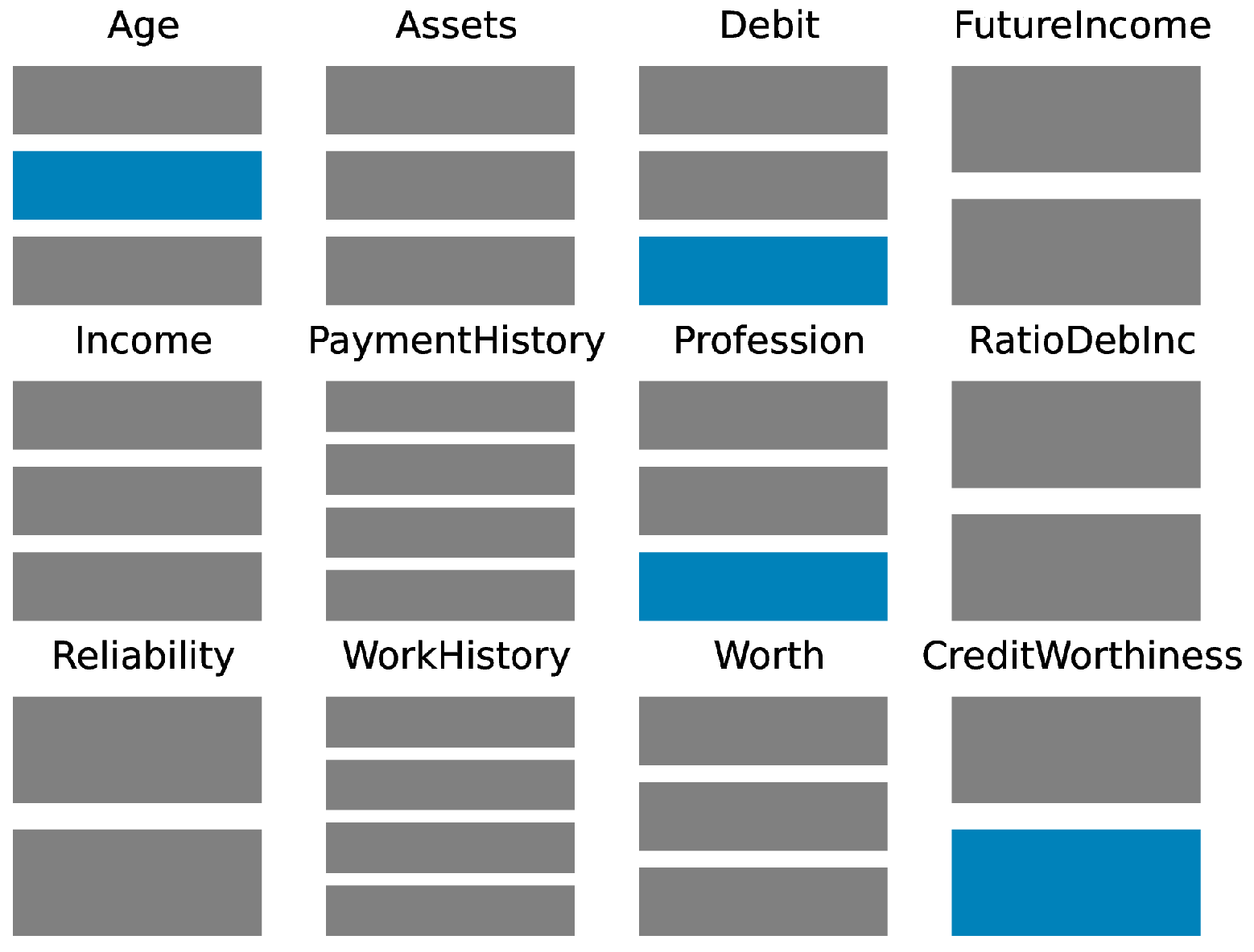} 
    \caption{\itr~\texttt{SAT} verification query result. The top figure represents the query. The two bottom figures illustrate counterexamples with feature values (blue) that fall within the constraints but lead to an undesired outcome. Feature values denoted in gray are irrelevant to the query when the blue values are present.}
    \label{fig:ITR_SAT}
\end{figure}

Next, we analyze the \itr query that evaluates to \texttt{SAT} for $M = BN4$. The first such query (in lexical order) corresponds to rule4 from Tab.\ref{tab:expert-rules}. Fig.\ref{fig:ITR_UNSAT} presents the verification query (top) along with two counterexamples (bottom) that demonstrate combinations of feature values leading to the undesired prediction outcome. For instance, a partial assignment including $Age = 0$ (or $Age = 1$ in the second counterexample), $Debit = 0$, and $Profession = 0$ results in a $Negative$ outcome for the $CreditWorthiness$ variable, thus invalidating the expected behavior.

\paragraph{\bf Verifying \fmo}
The process of performing \fmo verification is similar to \itr. Here, we provide a simplified overview of a single such query in practice. For our example, we define the following:

\smallskip

\partialassignment $= (x_{Worth} = 2, x_{Debit} = 0, x_{Profes.} = 1, x_{WrkHis.} = 3, x_{PaymHis.} = 3)$

\smallskip

$x_i = x_{Assets}$.

\smallskip

In human-readable terms, this definition corresponds to a person with high net worth, no debt, a medium-income profession, a stable work history, and an excellent payment history. This individual should receive a creditworthiness assessment that is monotonic with respect to their assets. Specifically, under the given partial assignment, we should never observe a situation where a person with $Assets = Poor$ or $Assets = Wealthy$ receives a $CreditWorthiness = Positive$ rating, while a person with $Assets = Average$ receives a $CreditWorthiness = Negative$ rating.

After running this \fmoN query on all models used in the previous verification task, we obtained the following results: ${B0, B1, B2, B3, B6, B8, B9} \in \texttt{UNSAT}$ and ${B4, B5, B7} \in \texttt{SAT}$. This result indicates that models B4, B5, and B7 do not adhere to the aforementioned query. For further analysis, counterexamples could be presented to experts with the goal of explaining which feature combinations lead to violations of this \fmoN.

\section{Conclusions}
In this work, we presented a SAT-based approach for verifying the adherence of Bayesian networks to specific design specifications. We introduced two verification queries aimed at assessing key behavioral aspects, facilitating efficient and effective testing of BN models prior to deployment—especially critical in safety-critical and high-stakes applications. Our method provides a practical solution for conducting both quick sanity checks and comprehensive verification procedures. By demonstrating its capability to handle reasonably sized BNs within a feasible timeframe, we established the viability of integrating our approach into a comprehensive testing suite. To underscore the real-world applicability of our framework, we showcased a case study illustrating its effectiveness in identifying potential errors within a real-world BN model. 

In future work, we aim to extend our approach to multi-class and multi-label BN classifiers, thereby broadening its applicability across various domains. Additionally, we plan to develop a model refinement framework that incorporates verification results, enabling retraining of models while accounting for the identified counterexamples.

\section{Acknowledgments} 
This work was partially supported by DARPA (award HR00112420329), U.S.\ Army (award W911NF-20-D0002), and by a Space Technology Research Institutes grant from NASA’s Space Technology Research Grants Program.

%
%
%
\bibliographystyle{splncs04}
\bibliography{references2}

\end{document}